\pdfoutput=1
\documentclass[11pt]{article}

\usepackage[final]{acl}
\usepackage{times,siunitx}
\usepackage{latexsym,lipsum}
\usepackage{booktabs}
\usepackage{multirow}
\usepackage{xcolor,colortbl}
\definecolor{darkgreen}{RGB}{0,120,0}
\definecolor{mediumgreen}{RGB}{120,200,120}
\definecolor{lightgreen}{RGB}{200,255,200}

\usepackage{tcolorbox} % For colored boxes

\usepackage[T1]{fontenc}
\usepackage[utf8]{inputenc}

\usepackage{microtype}
\usepackage{adjustbox} 

\usepackage{inconsolata}

\usepackage{graphicx}

\title{How Much is Too Much? Exploring LoRA Rank Trade-offs for Retaining Knowledge and Domain Robustness}
\author{Darshita Rathore\quad Vineet Kumar\quad Chetna Bansal \quad Anindya Moitra \\% ... \and Author n \\
        PayPal Artificial Intelligence\\PayPal, Bengaluru, India \\
        \texttt{\{drathore,vkumar32,cbansal,amoitra\}@paypal.com}
        }

\begin{document}
\maketitle
\begin{abstract}
Large language models are increasingly adapted to downstream tasks through fine-tuning. Full supervised fine-tuning (SFT) and parameter-efficient fine-tuning (PEFT) methods, such as Low-Rank Adaptation (LoRA), are two dominant approaches. While PEFT methods are widely used for their computational efficiency, the implications of their configurations (e.g., rank) remain under-explored in downstream Q\&A tasks and generalization. In this work, we perform a comprehensive evaluation across multiple reasoning and recall datasets, conducting a rank sweep to quantify the trade-off between SFT and PEFT. We also compare the accuracy of PEFT and SFT models across in-domain and out-of-domain adaptation, highlighting distinct generalization behavior and task-specific forgetting. We demonstrate that LoRA achieves competitive and in some cases superior performance compared to SFT, particularly on reasoning tasks at specific rank values. Additionally, we analyze the internal representations via spectral features and layer-wise attention structures, offering insights into representational drift and structural changes in attention patterns.
\end{abstract}

\section{Introduction}

Large Language Models (LLMs) have become indispensable for a wide range of use cases, including text generation, machine translation, summarization, question answering, data synthesis \& insights generation and software development, to name a few. Beyond these core tasks, LLMs are increasingly embedded in AI-powered agents for more complex, real-world workflows such as document understanding, data extraction, financial analysis, legal research, and web-based intelligence gathering \cite{genicious,minaee2025largelanguagemodelssurvey}. Their ability to operate across diverse domains with minimal supervision has led to rapid adoption in enterprise and production settings.

Despite their impressive capabilities, aligning LLMs with specific domains or use cases typically requires task adaptation via fine-tuning. Full supervised fine-tuning (SFT) -- where all model parameters are updated can improve performance, especially in high-stakes domains like law, finance, and medicine. However, SFT is computationally and memory intensive, often rendering it impractical at scale due to the size of modern models. Moreover, full fine-tuning poses challenges related to catastrophic forgetting \cite{haque2025catastrophicforgettingllmscomparative} and particularly when adapting to multiple tasks or clients in dynamic production environments.

Parameter-Efficient Fine-Tuning (PEFT) methods, most notably Low-Rank Adaptation (LoRA) \cite{hu2022lora}, have emerged as effective and scalable alternatives to full supervised fine-tuning. By injecting trainable low-rank matrices into the attention and feedforward layers of the model, LoRA enables fine-tuning with significantly fewer parameters, often without compromising task performance. This efficiency allows practitioners to maintain lightweight, domain-specific adapters while reusing a shared base model, thereby reducing both training and deployment costs.

While LoRA and other PEFT methods have been the focus of numerous empirical and theoretical investigations, most existing studies either emphasize absolute performance gains or analyze specific tasks in isolation. Few have systematically compared the structural and behavioral changes induced by LoRA with those resulting from full supervised fine-tuning (SFT) in a model-agnostic manner. However, a unified understanding of how different fine-tuning strategies, particularly LoRA with varying rank configurations, affect internal representations, generalization, and forgetting across reasoning and factual recall tasks remains lacking. In this work, we address this gap through a comprehensive evaluation framework that connects performance metrics with interpretability and model dynamics.

Building on this direction, we present a comprehensive study of LoRA rank selection and its effects on model behavior and performance. Our main contributions are as follows:

\begin{enumerate}
    \item We systematically evaluate how varying the LoRA rank affects downstream performance across multiple datasets and domain setups.

\item We compare SFT and PEFT approaches to assess whether full fine-tuning offers consistent benefits over parameter-efficient methods, especially for recall \& reasoning tasks.

\item We examine cross-domain performance degradation post fine-tuning, quantifying both forgetting and loss of generalization.

\item We analyze how internal representations, attention patterns, and layer-level drift differ from the base model after SFT and PEFT.

\end{enumerate}

\section{Related Works}

Full supervised fine-tuning (SFT) of large language models is a computationally expensive process. This has motivated extensive research on {parameter-efficient fine-tuning} (PEFT) techniques that adapt models without updating all weights. 

\citet{han2024parameter} provides a comprehensive taxonomy of PEFT approaches, categorizing them into four primary families: \textit{additive methods} (e.g., adapters, soft prompts), \textit{selective methods} (e.g., parameter masking), \textit{reparameterized methods} (e.g., low-rank decomposition), and \textit{hybrid approaches}. These methods substantially reduce memory and computational cost while maintaining downstream accuracy, enabling rapid task specialization even for billion-parameter models. Among these, low-rank reparameterization has emerged as a particularly compelling trade-off between efficiency and representational flexibility, allowing practitioners to inject compact task-specific capacity with minimal inference overhead.

Low-Rank Adaptation (LoRA) \citep{hu2022lora} injects trainable low-rank matrices $\Delta W = BA$ into existing linear projections and learns only $A, B$ while freezing the pretrained weights. The learned update can be merged into the base weights at inference time, introducing no additional latency. The original work demonstrated competitive accuracy with orders of magnitude fewer trainable parameters, but offered limited guidance on how to select the rank parameter across task families. 

Subsequent analyses by \citet{biderman2024lora} compared LoRA to full supervised fine-tuning and argued that LoRA ``learns less and forgets less'' highlighting that the effective update induced by SFT often possesses substantially higher intrinsic rank and thus greater capacity to both specialize and overwrite pretrained knowledge. \textit{Catastrophic forgetting} -- a phenomenon where a model trained on a new task drastically forgets previously learned information remains a central challenge in model adaptation \cite{haque2025catastrophicforgettingllmscomparative}. LoRA has been shown to mitigate forgetting on out-of-domain tasks, offering a favorable alternative. However, prior studies typically fixed the LoRA rank $r$ to a small constant, leaving open the quantitative relationship between $r$ and both in-domain and cross-domain behavior under-explored.

\citet{ren2024analyzing} further explored this issue and demonstrated that LoRA reduces forgetting compared to full fine-tuning, particularly when applied selectively. They introduced \textit{Interpolation-based LoRA} (I-LoRA), which leverages mode connectivity and a dual-memory learning mechanism to balance plasticity and stability. These approaches reveal that the geometry of the adaptation subspace strongly influences retention and transfer, motivating finer-grained analyses of how capacity constraints shape representational change.

Generalization under domain shift remains critical for achieving robust real-world performance. Complementary work on understanding \textit{where} fine-tuning alters model representations has provided deeper insight into PEFT behavior. \citet{hao2020investigating} used Jensen–Shannon divergence and Singular Vector Canonical Correlation Analysis (SVCCA) to show that BERT fine-tuning predominantly modifies upper layers, leaving lower layers largely intact across tasks. Such layer-wise analyses motivate studying how low-rank updates affect internal representations differently from full fine-tuning.

Our study builds upon these lines of research by systematically examining the \textit{rank–performance trade-off} in LoRA and its consequences for knowledge retention, reasoning ability, and cross-domain generalization. 

% TODO: Situate \cite{schulman2025lora}

\section{Methodology}

\subsection{Models and Fine-Tuning Configuration}
For our experiments, we used two instruction-tuned language models: \textbf{LLaMA-3.1-8B-Instruct} \cite{grattafiori2024llama3herdmodels} and \textbf{Qwen-2.5-7B-Instruct} \cite{qwen2025qwen25technicalreport}. Both models belong to the latest generation of open-source foundation models and demonstrate strong performance across a wide range of standard benchmarks. Their instruction-following capabilities and architectural improvements make them well-suited for evaluating diverse Q\&A tasks.

\subsection{Tasks and Datasets}
For our experiments, we considered three broad families of question-answering (Q\&A) tasks: general knowledge, mathematical reasoning, and domain-specific specialized tasks. These include both free-form text generation and multiple-choice question formats. Specifically, we used the GSM8K \cite{cobbe2021gsm8k}, MMLU \cite{hendrycks2021ethics}, and MedMCQA \cite{medmcqa} datasets to represent the three task categories. GSM8K focuses on grade-school level mathematical reasoning, MMLU (Massive Multitask Language Understanding) covers a wide range of general knowledge domains, and MedMCQA targets medical domain expertise. The data distribution across the datasets is as follows: GSM8K consists of 7,473 training and 1,319 test examples; MMLU includes 99,842 training and 14,042 test samples; MedMCQA contains 182,822 training and 6,150 test instances. Furthermore, we incorporated MathQA \cite{amini-etal-2019-mathqa} (with 29,837 training and 3,589 test samples) and LegalMCQ (940 training samples) to diversify the reasoning, domain-specific, and cross-domain evaluation. More details on data distribution are detailed in Section \ref{data-dist}.

\subsection{Evaluation Metrics}\label{eval-def}
All of our evaluation datasets consist of question-answer (Q\&A) formats. Except for GSM8K, all are structured as multiple-choice questions, which allows for reliable and consistent performance measurement through exact answer matching. This design choice was intentional: multiple-choice formats enable clear answer boundaries and reduce ambiguity in evaluating correctness. In case of GSM8K, where answers are free-form numeric responses with CoT style reasoning preceding it, we leverage the dataset’s annotation convention where the final answer is always prefixed with the token \texttt{\#\#\#}. This allows for straightforward extraction of the predicted answer using regular expressions. For all datasets, we consider a prediction correct if the generated answer string exactly matches the ground-truth answer string. We adopt accuracy (the proportion of correctly answered questions) as our primary evaluation metric.

\section{Experimental Setup}
\subsection{Performance Trade-offs}

To evaluate the performance of the base models, various LoRA configurations, and the full SFT models, we fine-tune each model on the training split of the respective datasets and evaluate them on the corresponding test sets\footnote{Except for {MedMCQA}, where ground truth answers are not available in the test set.}. 

Just with a small fraction of trainable parameters, LoRA achieves competitive downstream performance \cite{shuttleworth2025loravsfinetuningillusion}. 
% For example, \cite{ ding-etal-2025-sulora} outlines that a carefully designed variant of LoRA performs better than the standard LoRA for both accuracy and parameter efficiency. 

To evaluate the trade-off between model performance and fine-tuning, we train and assess each model in three configurations:
\begin{itemize}
    \item \textbf{Base model:} Original pre-trained \textit{off-the-shelf} base models were evaluated in a zero-shot setting using prompt-based inference, serving as a baseline.
    \item \textbf{LoRA fine-tuning:} LoRA configurations (except rank and alpha), including target modules (Key, Query, Value, and the Output layer), dropout, etc., were kept constant across all the experiments to experiments to enable a controlled and directly comparable evaluation.
    We sweep across five adaptation ranks $r \in \{8, 16, 32, 64, 128\}$ to analyse performance trends across varying levels of trainable parameter capacity. Setting $\alpha = 2\times r$ has been empirically shown to improve results~\cite{shuttleworth2025loravsfinetuningillusion} and avoid intruder dimensions with better generalization~\cite{biderman2024lora}.
    \item \textbf{Full-SFT:} Standard full supervised fine-tuning of all model weights, representing the upper bound in terms of adaptation flexibility and computational cost.
\end{itemize}
Further, hyperparameters such as the number of epochs, learning rate, optimizer, scheduler type, and maximum sequence length were kept constant during training. For inference, the parameters were matched to each model’s training configurations.

The accuracy metric used for evaluation is defined in Section~\ref{eval-def}, and the results are summarised in Table~\ref{tab:model_comparison}.

\begin{table*}[h!]
\centering
\resizebox{\textwidth}{!}{%
\begin{tabular}{@{}llccccccc@{}}
\toprule
\multirow{2}{*}{\textbf{Model}} & \multirow{2}{*}{\textbf{Dataset}} & \multirow{2}{*}{\textbf{Base Model}} & \multicolumn{5}{c}{\textbf{PEFT Model (Rank $r$)}} & \multirow{2}{*}{\textbf{Full SFT}} \\ \cmidrule(lr){4-8}
                                &                                   &                                      & \textbf{$r=8$} & \textbf{$r=16$} & \textbf{$r=32$} & \textbf{$r=64$} & \textbf{$r=128$} &                          \\ \midrule
\multirow{3}{*}{Llama-3.1-8B-Instruct} 
&  MMLU & 36.95\% & 57.39\% & 57.44\% & 57.21\% & 57.24\% & 57.20\% & 53.03\%\\
& GSM8K & 81.65\% & 65.35\% & 67.93\% & 69.83\% & 71.11\% & 70.43\% & 56.33\%\\
& MedMCQA  & 45.45\% & 51.90\% & 50.83\% & 51.67\% & 51.44\% & 51.67\% & 49.62\%\\\midrule
\multirow{3}{*}{Qwen-2.5-7B-Instruct} 
& MMLU & 31.15\% & 65.46\% & 65.87\% & 66.04\% & 66.15\% & 65.66\% & 60.90\%\\
& GSM8K  & 58.30\% & 66.94\% & 68.99\% & 71.80\% & 74.98\% & 70.05\% & 71.34\%\\
& MedMCQA  & 11.30\% & 27.24\% & 32.09\% & 21.40\% & 25.49\% & 21.17\% & 27.69\%\\ \bottomrule
\end{tabular}}

\caption{Model Performance Comparison: Base Model vs PEFT (LoRA Rank Sweep) vs Full SFT}
\label{tab:model_comparison}

\end{table*}

\subsection{Knowledge Retention, Forgetting \& Out-of-Domain Generalization}
Along with the task accuracy, we evaluated how much pre-trained knowledge is retained after fine-tuning for each model. Prior work has shown erosion of encoded world knowledge in language models with an increase in the amount of fine-tuning data~\cite{dou2024loramoealleviateworldknowledge}. To quantify this, we evaluate models both before and after fine-tuning on specific downstream tasks.

\begin{table}[b!]
\centering
\resizebox{0.5\textwidth}{!}{%
\begin{tabular}{@{}lllccll@{}}
\toprule
\textbf{Model} & \textbf{Trained on} & \textbf{Evaluated on} & \textbf{Base} & \textbf{LoRA} \\ 
\midrule

\multirow{5}{*}{\textbf{LLaMA}} 
& \multirow{3}{*}{MedMCQA} & LegalQA       & 34.25\% & 58.19\% \\
&                          & MathQA        & 31.20\% & 21.04\% \\
&                          & GSM8K         & 81.65\% & 74.37\% \\ 
& \multirow{2}{*}{GSM8K}   & MedMCQA       & 45.45\% & 46.51\% \\
&                          & Legal         & 34.25\% & 34.47\% \\

\midrule

\multirow{5}{*}{\textbf{Qwen}} 
% &                          & MathQA        & 25.97\% & 28.36\% \\
& \multirow{3}{*}{MedMCQA} & LegalQA       & 49.46\% & 63.94\% \\
&                          & MathQA        & 25.97\% & 28.36\% \\
&                          & GSM8K         & 58.30\% & 77.48\% \\
& \multirow{2}{*}{GSM8K}   & MedMCQA       & 11.30\% & 26.56\% \\
&                          & Legal         & 49.46\% & 57.02\% \\

\bottomrule
\end{tabular}}

\caption{Cross-task generalization performance of Base vs LoRA fine-tuned models on various QA datasets.}
\label{tab:cross_eval}
\end{table}

The experiments included:
\begin{itemize}
    \item As a proxy for factual retention, evaluating the model on knowledge-intensive tasks (e.g., MMLU).
    \item Comparing performance and generalization drop on benchmarks between LoRA and Full-SFT configurations.
    \item Evaluating on unseen domains (e.g., legal QA, math QA) with models fine-tuned on a specific task (e.g., MedMCQA).
\end{itemize}

The quantitative results from these experiments are presented in Tables~\ref{tab:model_comparison},~\ref{tab:cross_eval}, and~\ref{tab:ind_eval}. A detailed analysis and interpretation of these results, including model-wise and task-wise trends, is provided in the Discussion section (Section~\ref{eval-discussion}).

\begin{table}[b!]
\centering
\resizebox{0.5\textwidth}{!}{%
\begin{tabular}{@{}lllcc@{}}
\toprule
\textbf{Model} & \textbf{Trained on} & \textbf{Evaluated on} & \textbf{Base} & \textbf{LoRA} \\
\midrule
LLaMA & GSM8K & MathQA & 31.20\% & 22.32\% \\
Qwen  & GSM8K & MathQA & 25.97\% & 32.74\% \\
\bottomrule
\end{tabular}
}

\caption{Inter Domain generalization: Trained on a domain and evaluated on a similar domain but different distribution}
\label{tab:ind_eval}
\end{table}

\subsection{Training \& Inference Infrastructure}

Fine-tuning was performed on a compute cluster equipped with 4x NVIDIA H100 GPUs (80GB each) connected via NVLink, enabling high-throughput training for LLMs. We used mixed-precision training (bfloat16 where supported, otherwise fp16) to optimise GPU memory usage and computational speed. The \texttt{unsloth} \cite{unsloth} framework was employed for efficient LoRA fine-tuning with gradient checkpointing and support for large batch sizes via gradient accumulation. Distributed training was handled using HuggingFace’s \texttt{Trainer} \cite{vonwerra2022trl} API with PyTorch’s DDP backend.

For inference, we utilized \texttt{vLLM} \cite{vllm}, an optimised inference engine that supports paged attention and continuous batching, allowing for significantly faster and memory-efficient evaluation of the fine-tuned models. This setup enabled low-latency serving and efficient evaluation across multiple datasets and model variants.

\section{Discussion}\label{eval-discussion}
\subsection{Performance Trade-offs: LoRA's Efficiency and Efficacy}
The experimental results in Table \ref{tab:model_comparison} highlight LoRA's significant role as an effective and scalable alternative to full SFT. Across a majority of datasets and models, LoRA configurations consistently deliver substantial performance improvements over the base models, often exceeding those of Full SFT variants. This is consistent with the premise that PEFT methods can enable adaptation with significantly fewer parameters without compromising task performance.
A particularly noteworthy observation comes from the performance on the MMLU dataset. For both LLaMA-3.1-8B-Instruct and Qwen-2.5-7B-Instruct, LoRA configurations consistently outperform Full SFT. This outcome challenges the intuitive assumption that updating all model parameters through Full SFT would invariably lead to superior performance due to greater learning capacity. Instead, for general knowledge tasks like those in MMLU, which cover a wide range of domains, the constrained adaptation space of LoRA appears to act as a regularizer and by injecting only inherently low-rank matrices, LoRA limits the degrees of freedom for adaptation, potentially preventing the model from drastically altering its core knowledge base or overfitting to the specific training distribution. This preservation of broader pre-trained knowledge, coupled with targeted adaptation, led to better generalization and was more effective than a full fine-tune.

Findings of our experiments also highlight that there is \underline{no single rank that uniformly outperforms others}. However, the variability is minimal in certain classes of tasks like pure recall (MMLU \& MedMCQA) -- all ranks achieve almost similar performance; however, for more involved reasoning and math-based tasks, some ranks are better than others. This variability indicates the interplay between the nature, complexity of the task and data distribution in downstream performance. Tasks requiring more nuanced or extensive adaptations might benefit from a slightly higher rank, whereas simpler tasks or those where the base model already possesses strong foundational abilities might require less adaptation. 

Another interesting finding was with the GSM8K mathematical reasoning dataset. For LLaMA-3.1-8B-Instruct, both LoRA and SFT resulted in a significant decrease in accuracy compared to base model performance. This unexpected degradation suggests the latest model with good performance and strong instruction following capability may already exhibit superior performance, and SFT/PEFT is not beneficial. It is also likely that fine-tuning on specific tasks could induce biases which lead to a general loss of mathematical abilities. This underscores the importance of carefully evaluating the base model for downstream tasks, as fine-tuning may not always be required.

\begin{figure*}[ht!]
    \centering
    \includegraphics[width=1\linewidth]{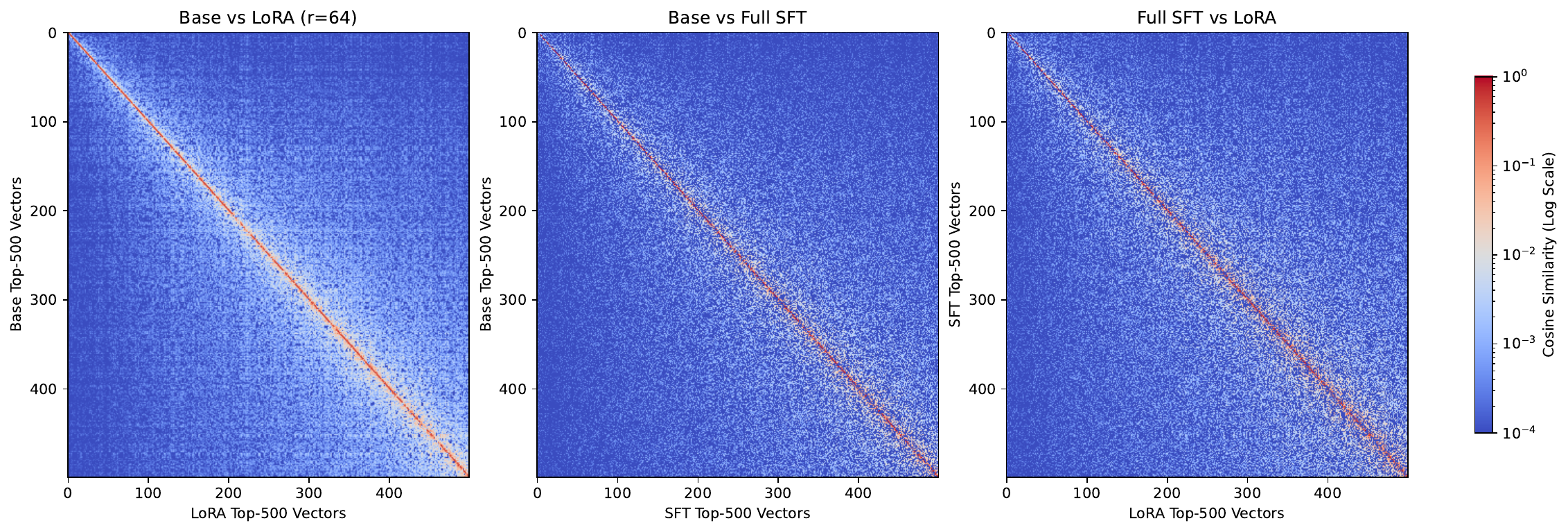}
    \includegraphics[width=1\linewidth]{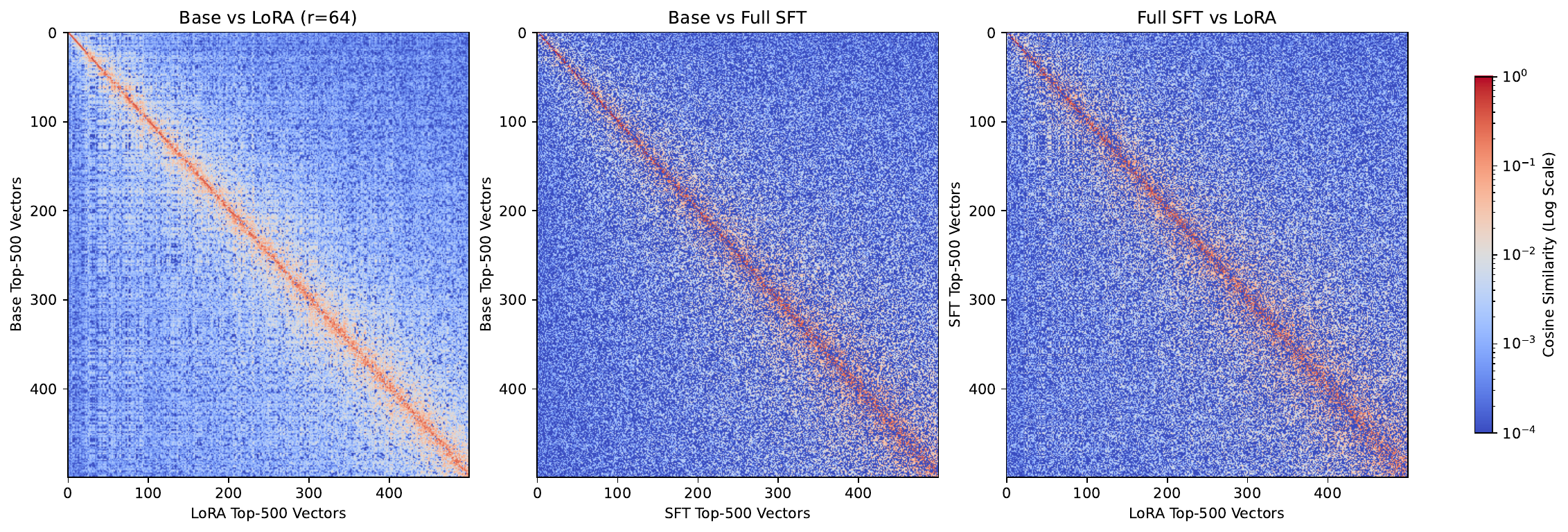}
    \caption{Similarity Heatmaps: Full SFT and for LoRA LLaMA GSM8K (top) and Qwen MMLU(bottom)}
    \label{fig:sim-heatmap}
\end{figure*}

\subsection{Generalization Capabilities: Cross \& Inter-Domain}
When models are trained on one domain and evaluated on another, LoRA-tuned models often exhibit robust generalization, sometimes even showing improvements over the base model. 
For example, Training on MedMCQA and evaluation on LegalQA, a significant performance improvement is observed (approximately 25\% and 15\% for LLaMA-3.1-8B-Instruct and Qwen-2.5-7B-Instruct, respectively). This suggests LoRA is not merely memorizing the facts specific to the training domain, rather it is learning more abstract and transferable skills. We conjecture this is true in the case of MedMCQA and LegalQA because both of these are factual recall-based tasks. However, this doesn't always hold in general; we also observed instances of negative transfer. For LLaMA-3.1-8B-Instruct, fine-tuning on MedMCQA led to a decrease in accuracy when evaluated on MathQA, dropping by $10\%$. This indicates a clear risk of catastrophic forgetting. The adaptation process for one task, particularly when the domains are fundamentally different (e.g., factual recall in medicine vs. reasoning in mathematics), might optimize the model's parameters in a way that conflicts with or overwrites internal representations crucial for other capabilities. 

Even within similar domains like mathematical reasoning, differences in data distribution between GSM8K and MathQA lead to LoRA adapters learning task-specific features that do not generalise to slightly different problem sets. Our experiments also highlight model-specific generalization behaviors. When trained on GSM8K and evaluated on MathQA, Qwen-2.5-7B-Instruct demonstrated an improvement in accuracy while LLaMA-3.1-8B-Instruct experienced a decrease. This divergence suggests that the underlying architectural differences between LLaMA-3.1-8B-Instruct and Qwen-2.5-7B-Instruct, such as specific attention mechanisms, normalisation layers, influence how effectively LoRA can adapt and generalise.

\subsection{Interpretability Analysis}

\paragraph{Spectral Features of Weight Matrices} We investigated how fine-tuning alters the fundamental characteristics of weight matrices by examining the similarity between their singular vectors before and after the fine-tuning process. 
We compute the cosine similarity between the top 500 singular vectors obtained via singular value decomposition (SVD) of the weight matrices to capture spectral shifts induced by adaptation. Figure \ref{fig:sim-heatmap} presents these similarity heatmaps for both LLaMA-3.1-8B-Instruct and Qwen-2.5-7B-Instruct models. The observed patterns indicate that the learning dynamics differ substantially between full supervised fine-tuning (SFT) and parameter-efficient fine-tuning (PEFT), suggesting distinct modes of representation change.
Full SFT, by modifying all model parameters, allows for a holistic and potentially more drastic reshaping of the entire representation space. While this can lead to superior optimisation for a specific task, it might also result in greater catastrophic forgetting of pre-trained knowledge. In contrast, LoRA, through the adaption of low-rank matrices into specific layers, induces more targeted changes and preserves the existing structure.

\begin{figure}[h!]
    \centering
    \includegraphics[width=\linewidth]{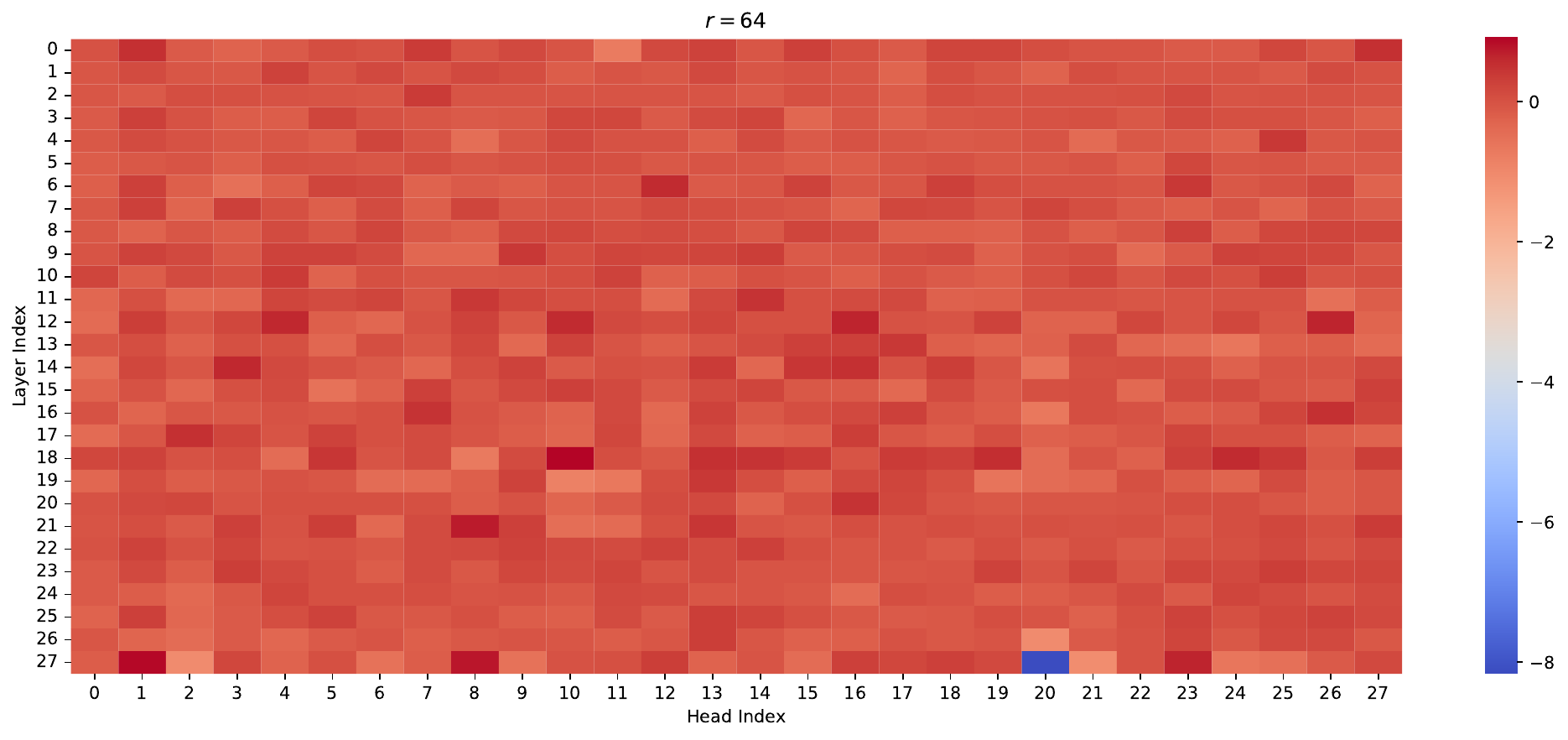}
    \includegraphics[width=\linewidth]{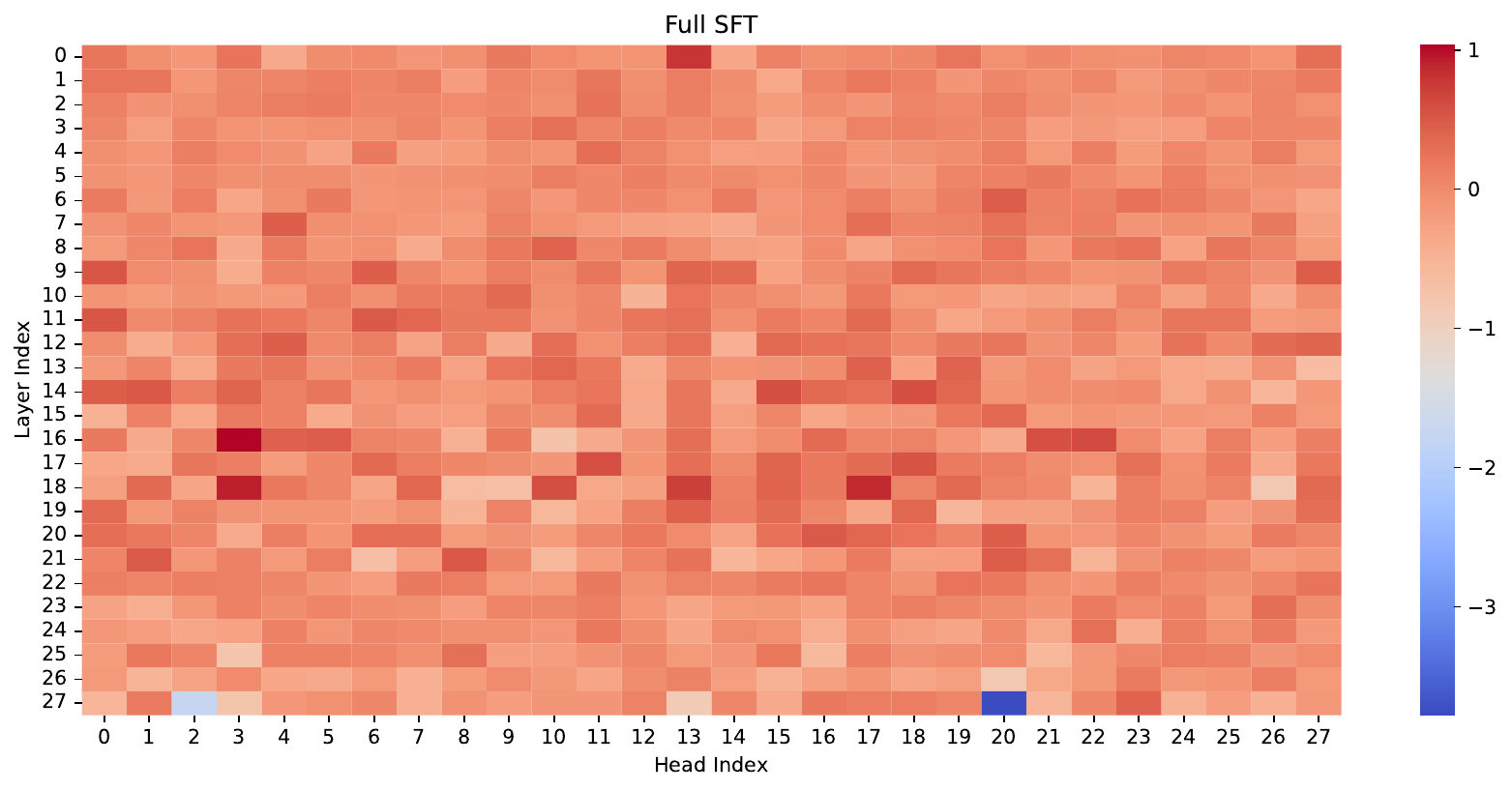}
    \caption{Log-Probability Drop After Attention Head Ablation for Qwen LoRA (top) and SFT (bottom)}
    \label{fig:qwen-attention-drop}
\end{figure}

\paragraph{Attention Head Ablation} As part of our interpretability analysis, we perform attention head ablation to identify which attention heads contribute most to the model’s output. For a given input, we systematically zero out individual attention heads and measure the drop in the log-probability of the correct answer. This is an established approach in many interoperability studies \cite{zhou2024role,michel2019sixteenheadsreallybetter}. A larger drop indicates that the head is more critical to the model’s prediction.
This approach allows us to quantify the functional importance of specific heads and track how this importance shifts across fine-tuning methods (LoRA vs. SFT) and task types (reasoning vs. recall). By comparing the ablation maps across models, we gain insight into how fine-tuning redistributes or reinforces focus on certain attention pathways. Figure \ref{fig:qwen-attention-drop} reveals that only a small subset of heads contribute significantly to task performance. The heatmaps show concentrated impact in mid-to-late layers, indicating that LoRA and SFT models rely on different attention pathways.

\paragraph{Frobenius ($l_2$) Norm}
% - Plot similarity matrices.

% - Deeper shifts in full fine-tuning; mid-rank(?) LoRA best retains base encoding

The Frobenius norm provides a proxy quantifying the overall magnitude of parameter changes during model adaptation. In the context of Low-Rank Adaptation (LoRA), this norm represents the cumulative strength of weight updates across the model, serving as a direct measure of how significantly the adaptation process modifies the base model's parameters. Formally, the Frobenius norm of a matrix is calculated as the square root of the sum of squared elements, effectively capturing the total magnitude of all parameter changes:

$$||\Delta W||_F = \sqrt{\sum_{i,j} (\Delta W_{ij})^2}$$

For LoRA adaptations, where weight updates are decomposed into low-rank matrices $(\Delta W = B\times A)$, the norm quantifies the effective strength of these adaptations while accounting for their interaction effects.

\begin{figure}[]
    \centering
    \includegraphics[width=\linewidth]{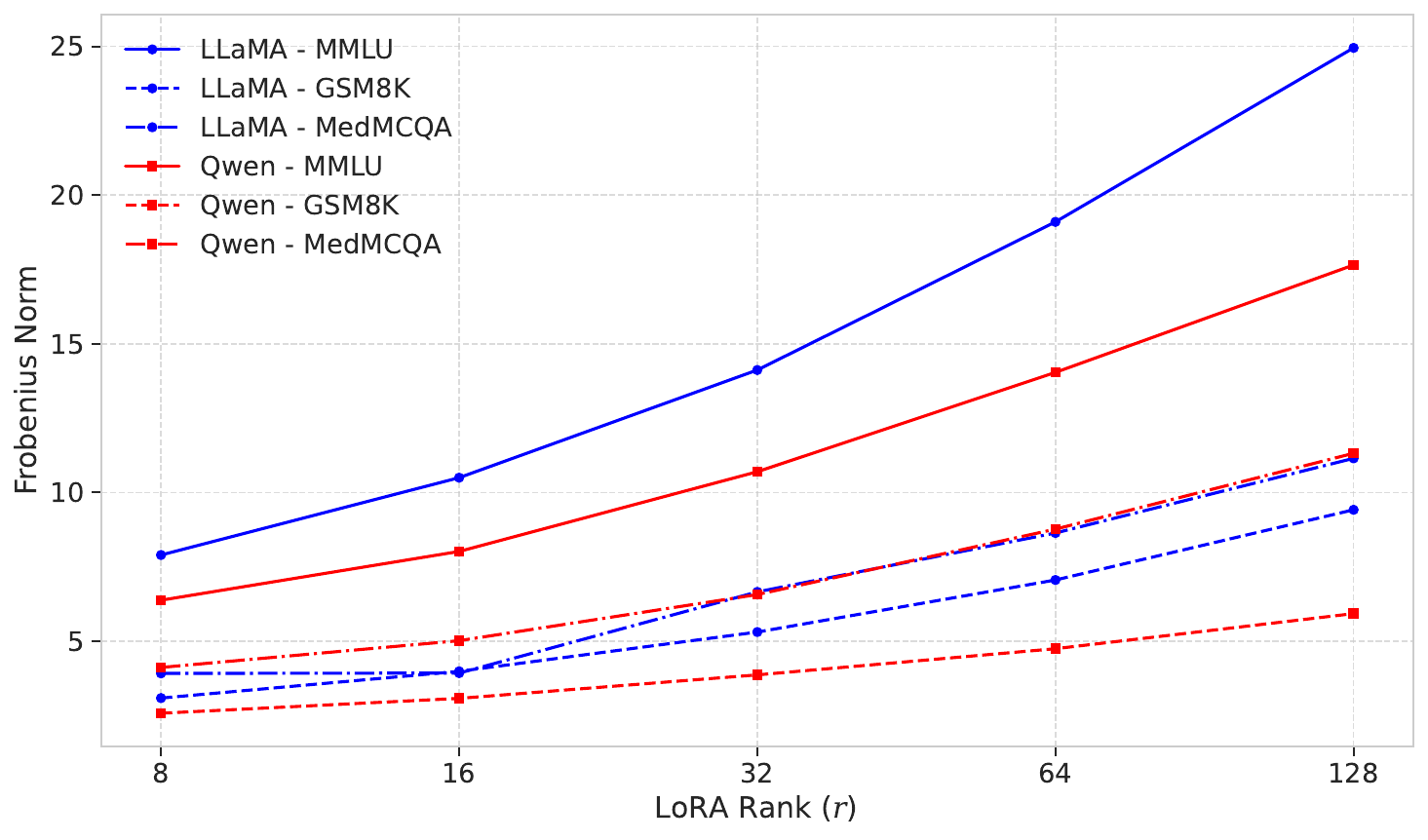}
    \caption{Quantifying Low-Rank Adaptation Impact: Frobenius Norm Scaling with Rank}
    \label{fig:fn}
\end{figure}

As evident in Figure \ref{fig:fn}, Frobenius norm exhibits approximately logarithmic growth with increasing LoRA rank across all model-task combinations.  MMLU adaptations consistently show the highest Frobenius norms, suggesting that general knowledge tasks require more substantial modifications.
GSM8K shows the lowest norm values for both models, indicating that mathematical reasoning capabilities might require more focused, rather than expansive, changes.

Looking at this from a model lens -- LLaMA-3.1-8B-Instruct models demonstrate steeper growth in Frobenius norm with increasing rank compared to  Qwen-2.5-7B-Instruct models, particularly for MMLU and GSM8K tasks, suggesting that LLaMA-3.1-8B-Instruct's architecture might be more responsive to increasing parametric capacity during adaptation.
Both models show similar patterns for MedMCQA adaptations, with Qwen-2.5-7B-Instruct exhibiting slightly higher norms at higher ranks, potentially indicating better alignment with medical domain adaptation.

\section{Conclusion}

This study advances a unified understanding of Parameter-Efficient Fine-Tuning (PEFT), specifically LoRA, by addressing gaps in how it performs across recall and reasoning tasks. Prior work often focused on isolated benchmarks, whereas we systematically evaluate how varying LoRA rank affects performance, generalization, and internal representations. 
% We show that LoRA can match or outperform full Supervised Fine-Tuning (SFT) in certain settings, with distinct structural changes such as spectral shifts and a logarithmic increase in Frobenius norm with rank. Our findings highlight how task and model characteristics influence LoRA's adaptation behavior, offering insights into its practical deployment.

Consistent with the \textit{no-free-lunch} principle \cite{mitchell1997machine}, there is no single universally optimal fine-tuning recipe for large language models. The effectiveness of adaptation depends on the interplay between task type, domain characteristics, and deployment constraints. Rather than proposing a one-size-fits-all rule, our contribution lies in establishing strong, evidence-based defaults that practitioners can reliably start from. Across reasoning and factual-recall tasks, we demostrate that LoRA provides a computationally efficient fine-tuning method that preserves general knowledge while maintaining competitive downstream performance. Empirically, intermediate ranks (\textit{r} = 32–64) offer a balanced operating point between representational capacity and stability, achieving robust performance.  We view these recommendations not as prescriptive choices, but as practical anchors that can be adapted to specific application contexts and model architectures.

\section*{Limitations}

This study provides a comprehensive analysis of LoRA and SFT; however, it is subject to certain limitations that also suggest avenues for future research. On LoRA configuration choices, the current methodology states that setting $\alpha =2\times r$ has been empirically shown to improve results and avoid `intruder dimensions’ \cite{shuttleworth2025loravsfinetuningillusion} with better generalization. This specific choice was applied consistently across experiments to ensure fairness. Future work could explore the impact of varying $\alpha$ independently of $r$ or investigate other LoRA variants and their respective optimal configurations.
This study focused on two specific instruction-tuned LLMs. Future work could extend this to a wider range of models (e.g., larger models, different architectures, non-instruction-tuned models) and compare LoRA against other PEFT variants (e.g., Prefix-tuning, Prompt-tuning, Adapter, QLoRA) to provide a more comprehensive understanding of the PEFT landscape. Furthermore, while diverse Q\&A tasks were covered, exploring other NLP tasks such as text generation, summarisation, or classification could yield additional insights into fine-tuning trade-offs. For deeper interpretability, future research could move beyond spectral features and attention ablation to explore other methods, such as neuron activation analysis, concept activation vectors, or causal mediation analysis, to gain a more granular understanding of how fine-tuning alters specific model behaviors. For observed negative transfer or performance degradation, future work could propose and evaluate mitigation strategies, such as multi-task fine-tuning with LoRA, selective LoRA application, or incorporating advanced regularization techniques.

\section*{Acknowledgments}
The authors thank the anonymous reviewers of AACL IJCNLP 2025 and OpenReview ARR for their insightful feedback and suggestions. The authors also dedicate this work to the memory of Prof. Pushpak Bhattacharya, whose vision, generosity, and teachings have inspired countless NLP researchers. 


\bibliography{custom}

\appendix
\section{Appendix}
\label{sec:appendix}
\subsection{Dataset Distribution}
\label{data-dist}
Table~\ref{tab:dataset} summarises the size and availability of splits across all datasets used in our experiments. MMLU serves as a general knowledge benchmark, while GSM8K and MATHQA target reasoning and numerical comprehension. MedMCQA and LegalMCQ cover domain-specific QA for the medical and legal domains, respectively.

\begin{table}[ht]
\centering
\caption{Dataset Statistics}
\resizebox{0.45\textwidth}{!}{%
\begin{tabular}{@{}l l r@{}}
\toprule
\textbf{Dataset} & \textbf{Split(s)} & \textbf{Number of Records} \\
\midrule
GSM8K    & train, test            & 7,473 / 1,319    \\
LegalMCQ & train                  & 940              \\
MATHQA   & train, dev, test       & 29,837 / 4,475 / 3,589 \\
MedMCQA  & train, validation, test& 182,822 / 4,183 / 6,150 \\
MMLU     & train, validation, dev, test & 99,842 / 1,531 / 285 / 14,042 \\
\bottomrule
\end{tabular}}
\label{tab:dataset}
\end{table}

% \begin{figure}
%     \centering
%     \includegraphics[width=1\linewidth]{latex/am.png}
%     \caption{Attention Map for LLama}
%     \label{fig:enter-label}
% \end{figure}

\subsection{SFT and PEFT Configuration}

\textbf{Full-SFT:} For full supervised fine-tuning, we train all model parameters using the \texttt{adamw\_8bit} optimizer with a learning rate of 5e-5 and weight decay of 0.01. Training is performed using a \texttt{per\_device\_train\_batch\_size} of 2 and \texttt{gradient\_accumulation\_steps} of 4, yielding an effective batch size of 8 per update step. We fine-tune for 3 epochs, using a linear learning rate scheduler with 10\% warmup steps. Mixed precision is enabled, automatically selecting between \texttt{fp16} and \texttt{bf16} based on hardware support.

Evaluation and checkpointing are conducted every 500 steps. Training logs and metrics are reported via TensorBoard. All models are trained with a maximum sequence length of 2048 tokens.

\vspace{1em}
\textbf{PEFT with LoRA via Unsloth:} We apply parameter-efficient fine-tuning using the \texttt{unsloth} framework \cite{unsloth}, which wraps HuggingFace’s PEFT and TRL libraries \cite{vonwerra2022trl}. We use LoRA with attention projection layers (\texttt{q\_proj}, \texttt{k\_proj}, \texttt{v\_proj}, \texttt{o\_proj}) as target modules - this matches the architecture of LLaMA-3 and Qwen models.

Key LoRA hyperparameters:
\begin{itemize}
  \item \texttt{r = 32}
  \item \texttt{lora\_alpha = 64}
  \item \texttt{lora\_dropout = 0.0}
  \item \texttt{bias = "none"}
\end{itemize}

Gradient checkpointing is enabled via \texttt{use\_gradient\_checkpointing = "unsloth"}, improving memory efficiency. We do not use any quantisation techniques (i.e. \texttt{loftq\_config = None}, \texttt{use\_rslora = False}).

For training, we use the \texttt{SFTTrainer} class provided by Unsloth with:
\begin{itemize}
  \item \texttt{max\_seq\_length = 2048}
  \item \texttt{dataset\_text\_field = "text"}
  \item \texttt{packing = False} (no input packing used)
  \item \texttt{dataset\_num\_proc = 2} (parallel preprocessing)
\end{itemize}

We use the same tokeniser, data splits, and stopping criteria across both SFT and LoRA runs for consistency.

\subsection{System Prompt for Training \& Evaluation}

To guide model behavior across tasks, we prepend task-specific system prompts during both training and evaluation. These prompts are designed to reflect the dataset domain and expected output style.

For example:
\begin{itemize}
    \item \textbf{MMLU (Knowledge)}: \texttt{"You are a helpful AI assistant that specializes in multiple-choice questions. Solve this MCQ and provide the correct option."}
    \item \textbf{GSM8K / MathQA (Reasoning)}: \texttt{"You are a math expert. Solve the problem step-by-step and return the final answer."}
    \item \textbf{MedMCQA}: \texttt{"You are a medical assistant. Carefully analyse the question and provide the correct option."}
    \item \textbf{LegalMCQ}: \texttt{"You are a legal expert. Read the question and choose the most accurate answer."}
\end{itemize}

% \begin{tcolorbox}[colback=green!5!white, colframe=green!60!black, title=Datasets]
% \textbf{LogiQA} \cite{liu2020logiqa} is constructed from the logical comprehension problems from publicly available questions of the National Civil Servants Examination of China, which are designed to test the civil servant candidates’ critical thinking and problem-solving. This dataset consists of 8,678 QA instances. (Train: 7376; Eval: 651; Test: 651)\\
% \end{tcolorbox}

All prompts are applied consistently across training and evaluation to ensure stable behavior and performance alignment.

\subsection{Layer-wise Norm Distribution}

To better understand how different LoRA configurations affect adaptation across the model layers, we visualise the layer-wise norm distributions of the LoRA weights for LLaMA fine-tuned on the MMLU dataset.

\begin{figure}[ht]
    \centering
    \includegraphics[width=0.95\linewidth]{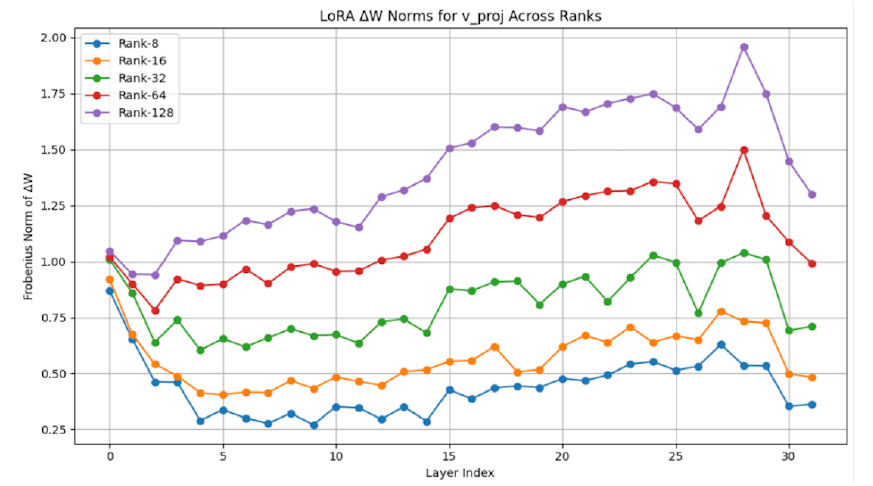}
    \includegraphics[width=0.95\linewidth]{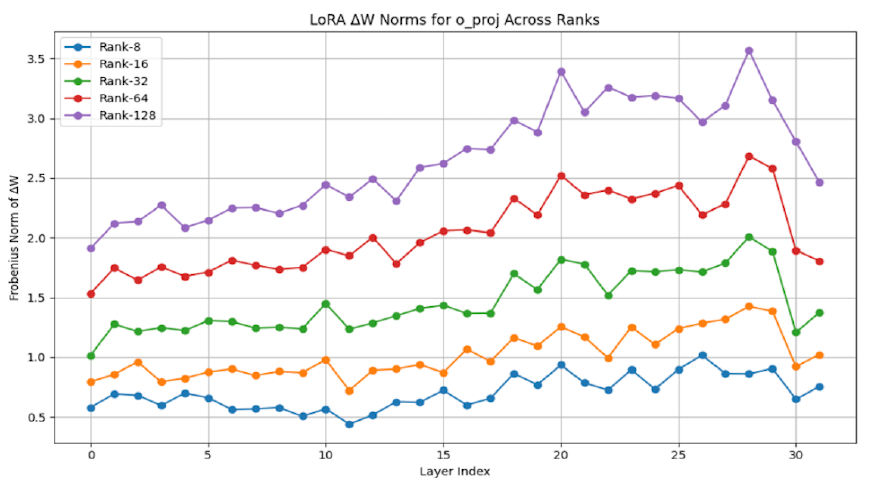}
    \includegraphics[width=0.95\linewidth]{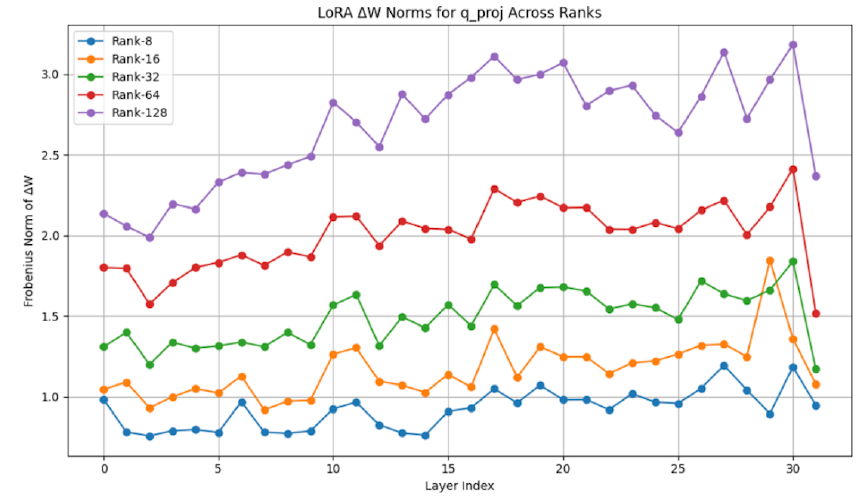}
    \includegraphics[width=0.95\linewidth]{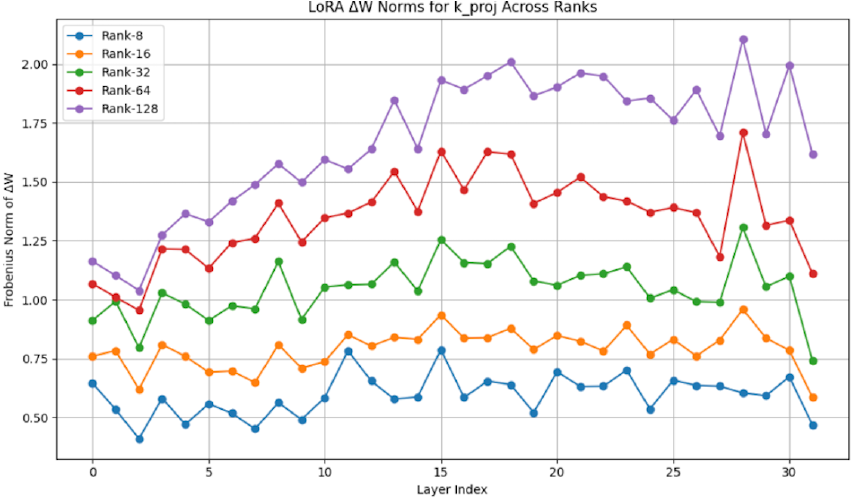}
    \vspace{-0.5em}

    \caption{Comparison of layer-wise LoRA adapter norm distributions across different ranks $r \in \{8, 16, 32, 64, 128\}$  for LLaMA fine-tuned on MMLU.}
    \label{fig:norm-all-ranks}
\end{figure}

Figure \ref{fig:norm-all-ranks} presents the $l_2$ norm of the injected LoRA deltas (adapter weights) across transformer layers for different LoRA ranks ($r$). These plots help identify which layers are more sensitive to adaptation and how this sensitivity varies with rank.

As observed, the middle and upper transformer blocks tend to accumulate more change as the rank increases, suggesting that LoRA adaptation is non-uniform across layers. This aligns with prior findings that later layers contribute more to task-specific reasoning and learning~\cite{hao2020investigating}. The layer norm trend can inform future decisions on layer selection for targeted PEFT.

% \begin{table*}[ht!]
% \centering
% \begin{tabular}{@{}llccccccc@{}}
% \toprule
% \multirow{2}{*}{\textbf{Model}} & \multirow{2}{*}{\textbf{Dataset}} & \multirow{2}{*}{\textbf{Base Model}} & \multicolumn{5}{c}{\textbf{PEFT Model (Rank $r$)}} & \multirow{2}{*}{\textbf{Full SFT}} \\ \cmidrule(lr){4-8}
% & & & \textbf{$r=8$} & \textbf{$r=16$} & \textbf{$r=32$} & \textbf{$r=64$} & \textbf{$r=128$} & \\ \midrule

% \multirow{3}{*}{Llama-3.1-8B-Instruct} 
% & MMLU & 36.95\% & \cellcolor{mediumgreen!60}57.39\% & \cellcolor{darkgreen!70}57.44\% & 57.21\% & 57.24\% & 57.20\% & 53.03\%\\
% & GSM8K & 81.65\% & 65.35\% & 67.93\% & \cellcolor{lightgreen!60}69.83\% & \cellcolor{darkgreen!70}71.11\% & \cellcolor{mediumgreen!60}70.43\% & 56.33\%\\
% & MedMCQA  & 45.45\% & \cellcolor{darkgreen!70}51.90\% & 50.83\% & 51.67\% & 51.44\% & \cellcolor{mediumgreen!60}51.67\% & \cellcolor{lightgreen!60}49.62\%\\ \midrule

% \multirow{3}{*}{Qwen-2.5-7B-Instruct} 
% & MMLU & 31.15\% & 65.46\% & 65.87\% & \cellcolor{mediumgreen!60}66.04\% & \cellcolor{darkgreen!70}66.15\% & \cellcolor{lightgreen!60}65.66\% & 60.90\%\\
% & GSM8K & 58.30\% & 66.94\% & \cellcolor{lightgreen!60}68.99\% & \cellcolor{mediumgreen!60}71.80\% & \cellcolor{darkgreen!70}74.98\% & 70.05\% & 71.34\%\\
% & MedMCQA & 11.30\% & 27.24\% & \cellcolor{darkgreen!70}32.09\% & 21.40\% & \cellcolor{mediumgreen!60}25.49\% & 21.17\% & \cellcolor{lightgreen!60}27.69\%\\ 
% \bottomrule
% \end{tabular}
% \caption{Performance comparison across PEFT ranks and Full SFT for two models. Top-3 per row highlighted in shades of green (dark = best).}
% \end{table*}

\end{document}